\documentclass[11pt]{amsart}
\usepackage{amsmath,amssymb,amscd}
\usepackage{amsmath,amssymb}
\usepackage{mathrsfs}
\usepackage{color}
\usepackage{amsxtra, amsmath}
\usepackage{amssymb, amscd}
\usepackage{graphicx}
\usepackage{caption}
\theoremstyle{plain}

\theoremstyle{definition}

\theoremstyle{remark}

\title[The relative importance of being Gaussian]{The relative importance of being Gaussian}

\author{F. Alberto  Gr\"unbaum \and Tongdi Xu}

\address{Department of Mathematics, University of California, Berkeley
CA 94720}
\email{grunbaum@math.berkeley.edu}

\address{School of Mathematical Sciences, Beijing Normal University, China} 
\email{202211130039@mail.bnu.edu.cn}



\begin{document}

\begin{abstract} 

	The remarkable results for denoising in computer vision using diffusion models given in \cite{SDWMG,HJA,HHG} yield a robust mathematical justification for algorithms based on crucial properties of a sequence of Gaussian independent $N(0,1)$ random variables. In particular the derivations use the fact that a Gaussian distribution is determined by its mean and variance and that the sum of two Gaussians is another Gaussian.

\bigskip

The issue raised in this short note is the following: suppose we use the algorithm without any changes but replace the nature of the noise and use, for instance, uniformly distributed noise or noise with a Beta distribution, or noise which is a random superposition of two Gaussians with very different variances. One could, of course, try to modify the algorithm keeping in mind the nature of the noise, but this is not what we do. Instead we study the performance of the algorithm when used with noise that is very far in nature from the Gaussian case, where it is designed to work well.

	Usually these algorithms are implemented on very powerful computers. Our experiments are all carried out on a small laptop and for the smallest possible image size. Exploring how our observations are confirmed or changed when dealing in different situations remains an interesting challenge.

\end{abstract}

\maketitle

\section{A brief  introduction}

The literature on the topic of denoising of images is huge and growing exponentially in time. In this brief note we concentrate on the remarkable results that originate with ideas presented in  \cite{SDWMG}, further developed in \cite{HJA} and analized with an applied mathematics audience in mind in \cite{HHG}.

\bigskip

Given the nature of this short note, it is impossible to give even a cursory review of the vast literature, and we refer the reader to the very short list of references given above, and their references. We just recall the fact that these ideas are inspired by an older and celebrated subject: non-equilibrium statistical mechanics. Needless to say this older subject has been the inspiration for very important work in machine learning. This was recognized with the $2024$ Nobel Prize in Physics awarded to J. Hopfield and G. Hinton.

\bigskip

A center piece of non-equilibrium statistical mechanics is the nonlinear integro-differential equation of L. Boltzmann, which describes convergence to a Maxwell-Boltzmann distribution in phase space and provides the backbone of the kinetic theory of gases. Several efforts to "derive" this equation by using probabilistic models have been proposed, in particular the idea of "propagation of chaos" put forward by Mark Kac \cite{MK}. One of us \cite{AG} dealt with this issue in his Ph.D. thesis (a long time ago) introducing the "duality viewpoint". In that paper a technical detail was left unresolved. Many years later this gap was filled by work that uses the "viewpoint" in \cite{AG} and is reported in \cite{MM,MMW} and \cite{CHD}. These papers contain many references to this line of work starting with L. Boltzmann. A very complete account of this circle of ideas, linking the classical limit theorems with the work of Boltzmann and more recent developments in Random matrix Theory is given in \cite{McK}.

\bigskip

\bigskip

\bigskip

\section{The contents of the paper}

\bigskip

\subsection{Experiment 1}
The first experiment implements a simplified 1D diffusion model to test how well a neural network can learn to reverse the diffusion process to generate samples close to the fixed target value $x0=7$ from different kinds of noise drawn from Gaussian, Uniform, or Beta($-0.5,-0.5$) distributions (each rescaled to have variance 1). We set the beta schedule to (0.0001, 0.02, 500). From the expression 
\begin{equation}
\mathbf{x}_t = \sqrt{\bar{\alpha}_t}\, \mathbf{x}_0 + \sqrt{1 - \bar{\alpha}_t}\, \boldsymbol{\epsilon}_t.
\label{eq:forward_diffusion}
\end{equation}
we can see that only$\sqrt{\bar{\alpha}_{500}}$ which is about $0.081$ of the information in $x0$ is retained, while the rest is noise. The neural network has one hidden layer with $32$ neurons and uses a learning rate of $1\times 10^{-3}$, with $3000$ epochs, $1000$ samples per epoch, and a batch size of $64$, enabling thorough training, stable optimization, and efficient computation.
We ran $100$ trials for each distribution (varying the choice of the seed), and the results are given below.
\begin{table}[h]
\centering
\begin{tabular}{lc}
\hline
Noise Type & Average of error \\
\hline
Beta($-0.5,-0.5$)     & 0.054226426 \\
Gaussian & 0.054758823 \\
Uniform  & 0.05340816 \\
\hline
\end{tabular}
\caption{Average error for different noise types.}
\label{tab:noise_error}
\end{table}

As we can see, the uniform distribution and the Beta($-0.5,-0.5$) distribution both perform as well as the Gaussian distribution yielding very good recovery.

\subsection{Experiment 2}
The second experiment aims to evaluate the model’s performance under a random mixture of noise conditions. The first distribution is the standard Gaussian distribution, noted as $N(0,1)$. The second distribution is a mixture with $0.9$ probability of $N(0,1)$ and $0.1$ probability of $N(0,100)$. The third distribution is a mixture with $0.5$ probability of $N(0,1)$ and $0.5$ probability of $N(0,100)$. We use these distributions in the forward process and then attempt to generate an image close to the original one. All other parameters are the same as in the first experiment in section $1$. The results are shown in Table $2$.

\begin{table}[h]
\centering
\begin{tabular}{lc}
\hline
Distribution         & Error        \\
\hline
com\_gaussian\_0.5 & 0.180548544 \\
com\_gaussian\_0.9 & 0.082517333 \\
gaussian            & 0.054758823 \\
\hline
\end{tabular}
\caption{Recovery errors for different noise distributions.}
\label{tab:noise_results}
\end{table}
From the results, we observe that as the noise distribution becomes more unstable, the reconstruction error increases. However, even with the $50$–$50$ mixture, the model is still able to achieve reasonably good recovey.

\subsection{Summary of Results}
Across these two experiments, we find that all distributions tested here can effectively replace the Gaussian distribution in the diffusion model while maintaining good performance. This is true whether the distribution is tightly clustered near the mean, uniformly spread, exhibits two peaks away from the mean, or represents a random mixture of two distinct distributions.

\bigskip

These results indicate that the neural network can achieve effective training and accurate generation even when the noise distribution does not satisfy the Gaussian assumptions required for the analytical validity of the diffusion model's transition equations.
As already mentioned in the abstract it would be of interest to explore these results further.

\section{Final comments}

After we finalized all the experiments reported here, with code written by one of us (T. X.), we came across \cite{BBCLKHGGG}, where considerations different but not unrelated to those here are raised. In our view our results and those in \cite{BBCLKHGGG} indicate that there may be room for further progress in this rich field.

\end{document}